# Multi-level Constrained Learning for Underwater Image Enhancement

Wangtao Bao, Liping Xie, *Member, IEEE,* Kanjian Zhang, and Huimin Lu, *Senior Member, IEEE*

***Abstract*—Underwater image enhancement aims to correct color and recover detail degraded by wavelength-dependent attenuation, scattering, and noise. Paired-reference supervision constrains output similarity, but does not by itself specify the color mapping, the strength of local detail corrections, or consistency with the degraded observation. We propose MCL-UIE, a paired-supervised framework that addresses these requirements by constraining restoration operations and the training objective. Monotonic Chromatic Transform (MCT) restricts global color mapping in log-LMS space through positive diagonal entries and bounded cross-channel coefficients. Uncertainty-Gated Wavelet Reconstruction (UGWR) modulates high-frequency residuals with a learned confidence gate calibrated against normalized wavelet reconstruction errors. During training, Reference-Anchored Degradation Replay (RADR) renders the enhanced image and its paired reference using shared latent degradation proxies, encouraging both to reproduce the input. The three components share a restoration network that combines structured color correction, locally controlled detail recovery, and reference-anchored replay. MCL-UIE ranks first in thirteen of sixteen full-reference dataset-metric comparisons and produces competitive visual results on no-reference benchmarks. Objective ablations indicate that confidence calibration is most effective when combined with replay supervision. RADR is used only during training and adds no inference-time computation. Our code is available at https://github.com/Wangtao-Bao/MCL-UIE.**

## I. Introduction

DUE to light absorption, scattering, and wavelength-selective attenuation in water, underwater images often suffer severe degradation, including color shifts, reduced contrast, uneven brightness, and loss of texture and structural detail. These degradations make underwater scenes difficult to analyze and interpret [1]. Underwater images are widely used in marine engineering, ecological monitoring, and detection by underwater robots [2], [3], [4]. Image enhancement can improve visibility for human observers and provide more reliable inputs for downstream tasks such as object detection and semantic segmentation [5], [6], [7], [8], [9]. Underwater image enhancement (UIE) thus supports both scene perception and the practical reliability of underwater vision systems.

Early UIE studies explored traditional image enhancement and model-based restoration [10], [11], [12], [13]. These methods improve color and contrast through image-domain processing or by estimating degradation parameters. However, handcrafted rules and assumed imaging conditions can limit their ability to adapt to different water types, lighting conditions, and scene content.

Deep learning has substantially advanced UIE [14], [15], [16], [17]. Trained on large datasets, these methods learn mappings from degraded observations to clear images and outperform traditional approaches in various underwater scenes. Their architectures, priors, and supervision terms differ, but supervision of the enhanced output alone does not specify how the network should transform color channels or regulate individual detail corrections. A low reconstruction error also does not establish whether the enhanced image can be re-degraded consistently with the input. These gaps motivate explicit constraints on intermediate restoration operations, used alongside output supervision.

Fig. 1 illustrates three considerations for controlling these operations. An unrestricted RGB mapping leaves the direction of channel transformations and cross-channel coupling unconstrained, which may destabilize image tones. High-frequency corrections may benefit from an explicit reliability estimate, as their responses can also contain noise, suspended particles, and weak scene edges. The enhanced result can also be evaluated for consistency with the degraded input, a relation that direct comparison with a paired reference does not test. These considerations concern global color transformation, local detail reconstruction, and the relation between enhancement and degradation, respectively. Accordingly, we consider constraints at three complementary levels: global chromatic transformation, local high-frequency reconstruction, and degradation consistency. Together, they provide structured control over different stages of the restoration process while retaining a shared reconstruction objective.

This work was supported in part by the National Natural Science Foundation of China under Grant 62372104, in part by the Fundamental Research Funds for the Central Universities under Grant 2242026RCB0009, and in part by the Research Fund of Key Program for Advanced Ocean Institute of Southeast University under Grant KP202402. Besides, we thank the Big Data Computing Center of Southeast University for providing the facility support on the numerical calculations. (Corresponding author: Liping Xie).

Wangtao Bao, Liping Xie, Kanjian Zhang, and Huimin Lu are with the Key Laboratory of Measurement and Control of Complex Systems of Engineering, Ministry of Education, School of Automation, Southeast University, Nanjing 210096, China; and also with Advanced Ocean Institute of Southeast University, Nantong 226010, China. (e-mail: wtbao@seu.edu.cn, lpxie@seu.edu.cn, kjzhang@seu.edu.cn, luhuimin@ericlab.org).

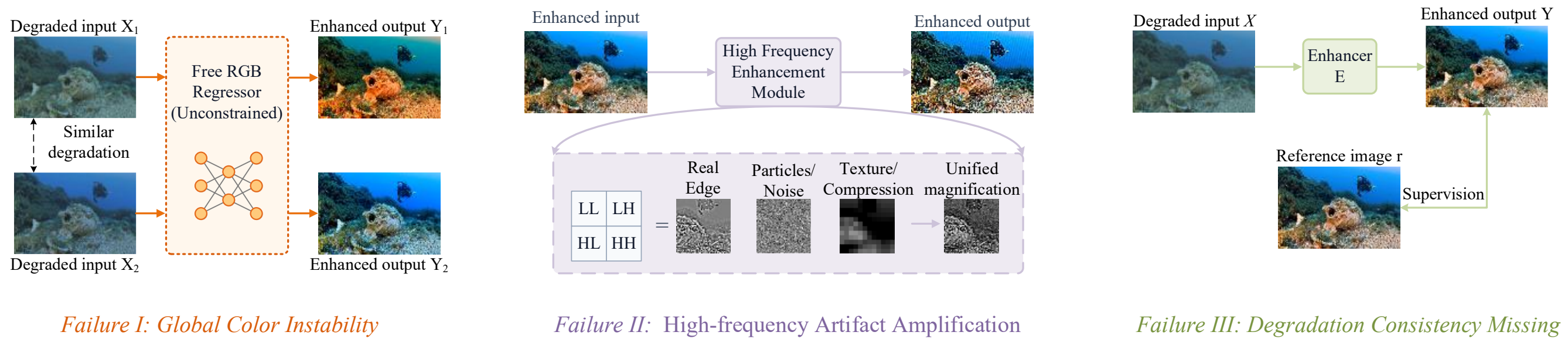

Fig. 1. Illustration of the challenges motivating MCL-UIE: unconstrained global color mapping, unreliable high-frequency enhancement that may amplify noise or artifacts, and ambiguous interpretation of degradation causes under paired supervision.

We propose Multi-level Constrained Learning for Underwater Image Enhancement (MCL-UIE), which combines these constraints in one restoration pipeline. Monotonic Chromatic Transform (MCT) first predicts a restricted global log-LMS mapping from robust image-level statistics and supplies the color-corrected input for spatial restoration. Uncertainty-Gated Wavelet Reconstruction (UGWR) predicts wavelet detail residuals and controls their magnitude through confidence calibrated against normalized reconstruction errors. During training, Reference-Anchored Degradation Replay (RADR) applies a shared renderer to the enhanced image and its paired reference, encouraging both paths to reproduce the degraded input. MCT and UGWR constrain restoration operations directly, while RADR constrains the output through a shared degradation model. A common backbone provides spatial features for reconstruction and degradation-proxy estimation. We train the three components jointly and evaluate their contributions through progressive component and training-objective ablations.

The main contributions are as follows:

1. We propose MCL-UIE to extend supervision beyond output fidelity by regulating key restoration operations. MCT, UGWR, and RADR respectively target global color transformation, local high-frequency correction, and degradation consistency.

2. We design MCT to predict a lower-triangular transform from robust log-LMS statistics, with positive diagonal entries and bounded cross-channel coefficients. For fixed transform parameters, this parameterization preserves positive self-channel responses in the linear log-LMS core and limits cross-channel influence.

3. We develop UGWR to regulate wavelet detail residuals with a bounded confidence gate. Calibration against detached, normalized reconstruction errors gives the gate an explicit local reliability target for controlling high-frequency corrections.

4. We introduce RADR, which re-renders the enhanced image and its reference during training using shared wavelength-dependent attenuation, backscattering, and distance proxies. Reference replay anchors the degradation parameters to the paired observation; prediction replay couples the enhanced output to those same parameters.

## II. Related Work

### *A. Traditional and Physics-Based UIE Methods*

Traditional UIE methods either enhance images without a physical model or restore them using a physical model. The first group uses white balance, color compensation, histogram adjustment, Retinex decomposition, or multi-scale fusion to improve color distribution and local contrast directly [11], [18], [19]. These methods generally require less computation because they do not estimate scene depth or water optical parameters, and they can produce relatively natural results in some scenes. However, their reliance on statistical priors or empirical rules makes it difficult to distinguish color attenuation, nonuniform illumination, and backscattering during light propagation through water. Fixed enhancement strategies can consequently overcompensate colors, overexpose local regions, or amplify noise when water type, lighting, or imaging distance changes substantially.

Physical model-based methods describe underwater light propagation and recover clear images by estimating transmittance, background light, attenuation coefficients, and scene distance. Early approaches adapted atmospheric scattering models, using dark-channel, red-channel, color-line, or wavelength-attenuation priors to estimate degradation parameters [20], [21], [22]. These methods offer physical interpretability, but their restoration quality depends strongly on the underlying assumptions and parameter estimates. Coupling between transmittance and background light can bias the estimates, particularly under artificial lighting, in nonuniform water, or in low-light conditions.

### *B. Deep Learning-Based UIE Methods*

Deep learning-based UIE has evolved from convolutional restoration to diverse representations and learning strategies. Early methods include UWCNN [14], which learns residual mappings from synthesized underwater images, and WaterNet [15], which fuses differently preprocessed inputs. FUnIE-GAN [23] introduces adversarial and perceptual constraints, while domain-adaptive UIE reduces distribution discrepancies across underwater environments [24]. UIEC$^2$-Net [25] and Ucolor [26] exploit complementary color spaces, whereas PUIE-Net [27] models enhancement through conditional distribution learning and consensus estimation.

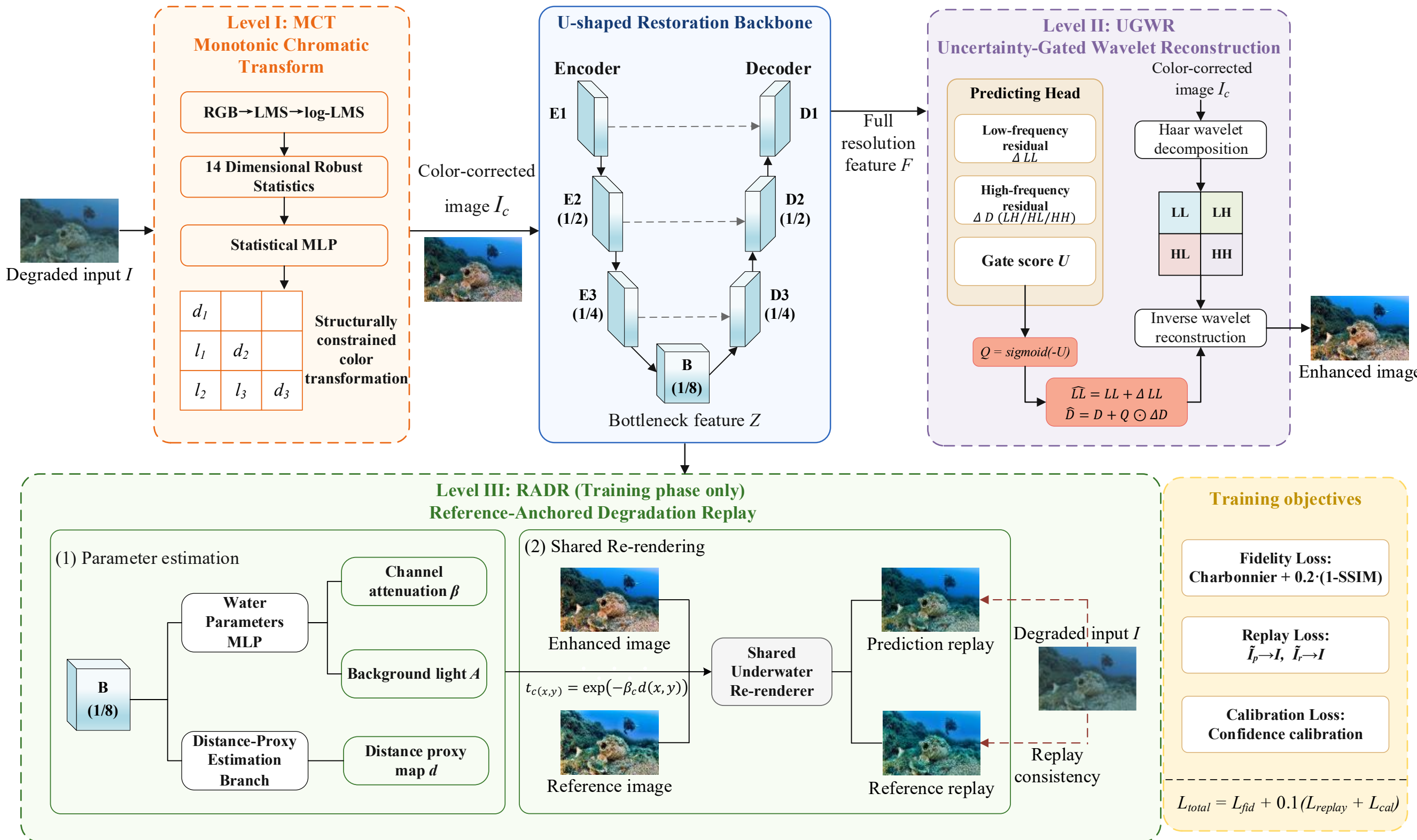

Fig. 2. Overall architecture of the proposed MCL-UIE framework. The framework comprises Monotonic Chromatic Transform (MCT), a U-shaped spatial restoration backbone, Uncertainty-Gated Wavelet Reconstruction (UGWR), and Reference-Anchored Degradation Replay (RADR). RADR is used only during training and incurs no computational overhead during inference.

Recent studies further explore degradation-aware representations and improved supervision. SS-UIE [28] combines spatial selective scanning with spectral attention and frequency-wise supervision. PSPL [29] addresses imperfect pseudo-references through quality-guided comparisons and contrastive learning. IA2U [30] adapts pretrained in-air models using degradation-related priors, while PGMamba [31] incorporates underwater imaging priors into state-space modeling.

These methods advance feature representation and supervision, but output fidelity alone does not explicitly regulate global color mappings, individual high-frequency corrections, or consistency with the degraded observation. MCL-UIE addresses these complementary aspects through MCT, UGWR, and RADR, respectively, introducing constraints on restoration operations and degradation consistency alongside paired-reference supervision.

### *C. Frequency-Domain Image Restoration*

Spatial-domain networks effectively extract local features but lack explicit frequency separation. Frequency-domain methods address this limitation by decomposing and integrating complementary representations. Wavelet-based Dual-stream Networks [32] separately process low- and high-frequency components, while UHD-SFNet [33] combines spatial and Fourier features for color and texture restoration. Subsequent methods strengthen cross-domain interactions: SFGNet [34] employs gradient-aware spatial-frequency fusion, FDCE-Net [35] integrates frequency-spatial residual modeling with dual-color representations, and CDF-UIE [36] explores cross-domain feature interaction. More recent approaches incorporate wavelet modeling into advanced architectures. USWformer [37] introduces sparse wavelet attention, WWE-UIE [38] combines wavelet enhancement with white balancing and gradient-aware gating, and WaveletFormer [39] integrates learnable wavelet filters with multi-scale Transformers.

Despite these advances, high-frequency components contain both genuine scene details and degradation-induced responses, making the reliability of individual corrections important. UGWR addresses this issue by adaptively controlling predicted wavelet residuals through confidence gates calibrated against normalized reconstruction errors. This design explicitly regulates local corrections across color channels and wavelet directions, complementing existing frequency representation and fusion strategies.

## III. Method

### *A. Overall Architecture*

Fig. 2 shows the MCL-UIE architecture. MCT applies a global chromatic transform, after which a U-shaped backbone extracts spatial features. UGWR uses these features to predict wavelet residuals and control high-frequency updates, yielding the enhanced image. During training, RADR predicts degradation parameters from the backbone bottleneck and uses the same parameters in the enhanced-image and reference-image replay paths. The network directly constrains the input color mapping and final detail corrections, while the training objective evaluates the output against both its paired reference and the degraded observation.

Given a degraded underwater image

$$I \in [0,1]^{3\times H\times W}. \quad (1)$$

MCT first corrects the image-level color. The spatial backbone then extracts full-resolution restoration features and bottleneck features, and UGWR uses the restoration features to reconstruct the image in the frequency domain:

$$\begin{cases} I_c = \mathcal{M}(I), \\ (F,Z) = \mathcal{B}(I_c), \\ \hat{J} = \mathcal{U}(I_c, F). \end{cases} \quad (2)$$

Here, $I_c$ denotes the color-corrected MCT output, $F$ and $Z$ denote the full-resolution and bottleneck features, respectively, and $\hat{J}$ is the final enhanced image.

During paired training, RADR estimates a shared set of latent degradation variables from $Z$ and uses them to re-render the enhanced image $\hat{J}$ and reference image $J$ in the degradation domain:

$$\begin{cases} \theta_d = g(Z) = \{\beta, A, d\}, \\ \tilde{I}_p = \mathcal{R}(\hat{J}; \theta_d), \\ \tilde{I}_r = \mathcal{R}(J; \theta_d). \end{cases} \quad (3)$$

MCT controls the global color mapping supplied to the backbone, UGWR controls the local corrections that form the output, and RADR constrains the relation between that output and the observation. We jointly optimize all three components under an objective that includes paired-reference fidelity. At inference, we retain only the restoration path in Eq. (2), so the replay branch adds no computation.

### ***B.*** *Monotonic Chromatic Transform*

Underwater color distortion often appears as a pronounced global imbalance between channels. Convolutional and attention networks can learn flexible RGB corrections, but need not restrict the direction or strength of cross-channel coupling. This flexibility may lead to overcompensation or tone shifts as water conditions change. MCT applies a restricted global transform before spatial restoration, reducing the need for the backbone to learn the entire color mapping implicitly. Image-level statistics determine the transform parameters. The parameterization controls self-channel responses and cross-channel influence. The backbone then reconstructs spatial content from the color-corrected input.

#### *1) Statistical Representation*

We map the input RGB image to LMS space and take its logarithm:

$$L = log[max(M_{RGB\to LMS} \times I, 10^{-4})]. \quad (4)$$

The fixed matrix $M_{RGB\to LMS}$ performs the RGB-to-LMS transformation. Taking the logarithm converts multiplicative changes in channel intensities into additive changes that are easier to model.

MCT uses a 14-dimensional statistical descriptor as input to the parameter predictor in place of high-dimensional image features:

$$s = [q_{0.05}, q_{0.50}, q_{0.95}, \mu_{RGB}, \bar{s}, \bar{g}] \in \mathbb{R}^{14}. \quad (5)$$

where $q_{0.05}$, $q_{0.50}$, and $q_{0.95} \in \mathbb{R}^3$ are the 5%, 50%, and 95% quantiles of the three log-LMS channels, respectively; $\mu_{RGB} \in \mathrm{R}^3$ contains the RGB-channel means; and $\bar{s}$ and $\bar{g}$ are the mean saturation and mean brightness gradient. The quantiles describe each channel's distribution range more robustly than a single mean, reducing sensitivity to isolated bright or dark regions. Together, the nine quantile statistics, three RGB means, and two global statistics form the descriptor. A lightweight statistical MLP maps this descriptor to the nine parameters of the constrained chromatic transform:

$$\begin{cases} p = \mathcal{F}_{stat}(s) \in \mathbb{R}^9, \\ \mathcal{F}_{stat}: \mathbb{R}^{14} \to \mathbb{R}^{64} \to \mathbb{R}^{64} \to \mathbb{R}^9. \end{cases} \quad (6)$$

#### *2) Constrained Chromatic Mapping*

To constrain the 3×3 transform, MCT splits the predicted vector p into three diagonal elements, three cross-channel elements, and three biases, with separate constraints for each group:

$$\begin{cases} d_i = Softplus(p_i), \\ l_i = 0.15tanh(p_{i+3}), \\ b_i = 0.25tanh(p_{i+6}). \end{cases} \quad (7)$$

For $i = 1,2,3$, these parameters form the lower-triangular color transform

$$T = \begin{bmatrix} d_1 & 0 & 0 \\ l_1 & d_2 & 0 \\ l_2 & l_3 & d_3 \end{bmatrix}, d_i > 0, |l_i| \le 0.15. \quad (8)$$

This parameterization bounds cross-channel compensation and restricts arbitrary coupling. With the transform parameters held fixed, positive diagonal entries ensure positive self-channel derivatives in the linear log-LMS core and a nonsingular transformation matrix:

$$det(T) = d_1 d_2 d_3 > 0, \quad (9)$$

We apply the transform in log-LMS space and clip the resulting RGB output to [0,1]:

$$\begin{cases} L_c = TL + b, \\ I_c = clip_{[0,1]}\left(M_{LMS\to RGB} exp\left[clip_{[-9,4]}(L_c)\right]\right). \end{cases} \quad (10)$$

Here, $b = (b_1, b_2, b_3)^T$ is shared across spatial locations, and clipping is element-wise. The fixed RGB-to-LMS matrix operates on normalized RGB values without inverse-gamma correction; $M_{LMS\to RGB}$ is its numerical inverse. We initialize the final MLP layer to produce an identity transform and zero bias, so MCT starts with minimal changes to the input colors before learning image-dependent corrections. Monotonicity applies to the forward self-channel response of the linear core in log-LMS space when its parameters are held fixed.

### ***C.*** *U-Shaped Spatial Restoration Backbone*

MCT primarily addresses global color imbalance and does not explicitly model spatial restoration associated with scattering and blur. A compact U-shaped encoder-decoder, with three downsampling stages and four spatial scales, handles spatial restoration. It extracts multi-scale features from the color-corrected image and produces full-resolution features for UGWR and bottleneck features for RADR.

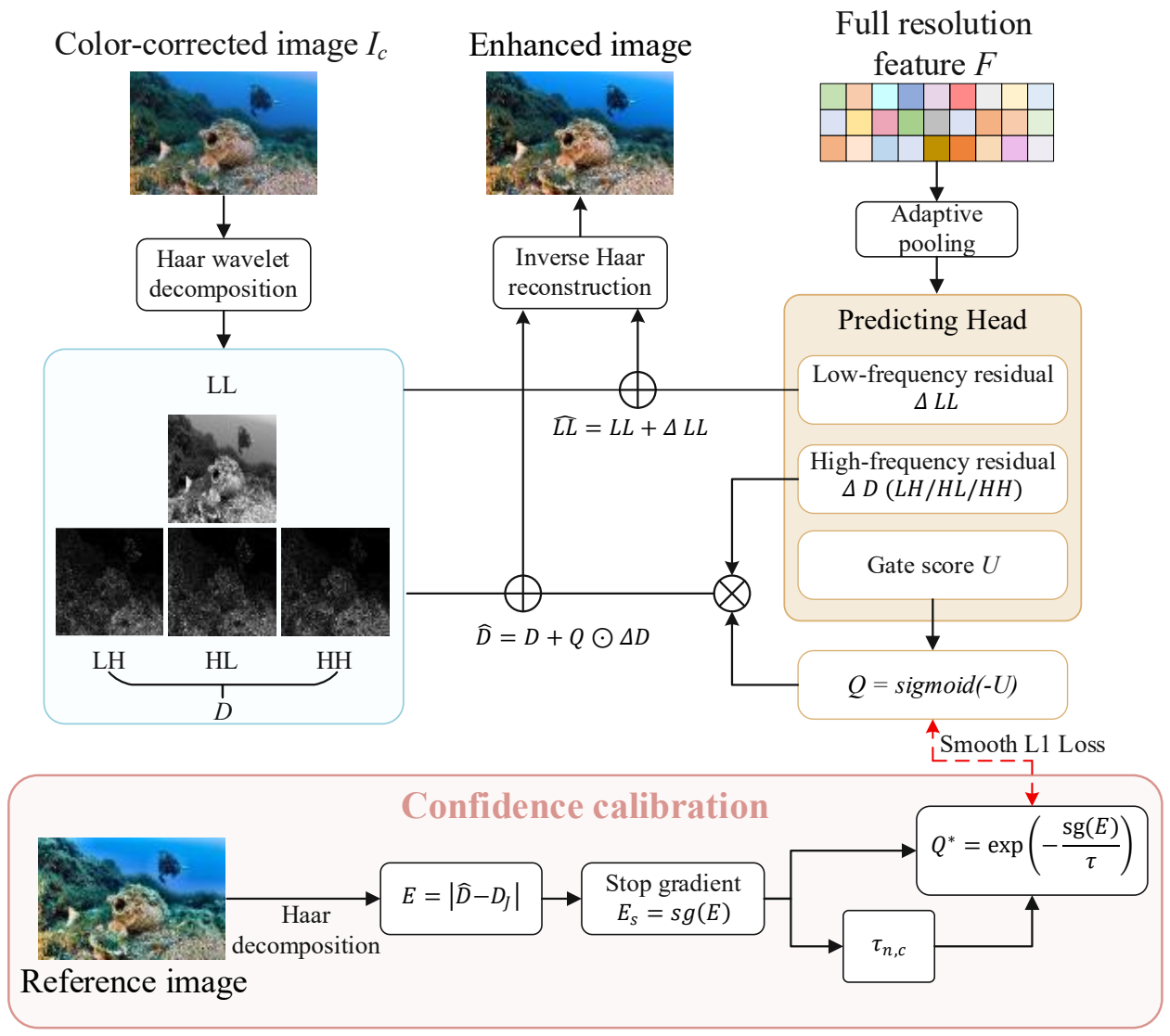

Fig. 3. Structure of the UGWR module. The color-corrected image $I_c$ is decomposed into low- and high-frequency wavelet subbands, while the pooled full-resolution feature $F$ predicts low-frequency residuals, high-frequency residuals, and confidence scores. The confidence gates high-frequency residual corrections and is calibrated during training using reference-detail errors.

The default encoder channel widths are $C$, $2C$, $4C$ and $8C$, with $C = 32$. Their spatial resolutions are

$$H \times W, \frac{H}{2} \times \frac{W}{2}, \frac{H}{4} \times \frac{W}{4}, \frac{H}{8} \times \frac{W}{8}. \tag{11}$$

Each stage comprises Residual Depthwise Blocks. For an input feature $X$, a restoration block computes

$$Y = X + \alpha \odot P\left[\phi\left(Dw\left(\phi\left(E_{\exp(N(X))}\right)\right)\right)\right]. \tag{12}$$

where $Y$ is the output feature, and $\odot$ denotes element-wise multiplication. $N(\cdot)$ denotes Group Normalization; $E_exp(\cdot)$, $Dw(\cdot)$, and $P(\cdot)$ denote the 1×1 channel expansion, 3×3 depthwise convolution, and 1×1 channel projection, respectively; $\phi(\cdot)$ is GELU; and α is a learnable per-channel residual scale initialized to 0.1.

The encoder produces multi-scale features $E_1$, $E_2$, and $E_3$ and the bottleneck feature $Z$. Let $D_4 = Z$. For $i = 3,2,1$, the decoder upsamples the lower-resolution feature to the corresponding scale, concatenates it with encoder feature $E_i$, and applies a 1×1 convolution followed by a residual restoration stage:

$$D_i = \mathcal{B}_i(\mathrm{Conv}_{1\times1}[Up(D_{i+1}), E_i]). \tag{13}$$

$Up(\cdot)$ denotes bilinear upsampling followed by a 3×3 convolution, whereas $\mathcal{B}_i$ denotes the residual restoration stage at scale $i$.

The backbone therefore produces two features with distinct roles:

$$F \in \mathbb{R}^{C\times H\times W}, \quad Z \in \mathbb{R}^{8C\times\frac{H}{8}\times\frac{W}{8}}. \tag{14}$$

UGWR uses the full-resolution feature $F$ to predict wavelet corrections, while RADR estimates shared degradation parameters from the bottleneck feature $Z$. Both spatial recovery and degradation estimation therefore use the same backbone, without an additional encoder.

### *D. Uncertainty-Gated Wavelet Reconstruction*

High-frequency components in underwater images contain both scene details and responses to noise or suspended particles. UGWR predicts candidate wavelet corrections, then controls the fraction applied at each spatial location and wavelet direction (Fig. 3). The learned gate is calibrated against normalized detail-coefficient reconstruction errors, defining confidence in terms of local reconstruction quality. The network thus separates candidate prediction from magnitude control, while training both operations jointly.

#### *1) Wavelet Decomposition*

A one-level reversible Haar transform decomposes the MCT output $I_c$ into one low-frequency and three directional high-frequency subbands. For each color channel, a, b, c, and d denote the top-left, top-right, bottom-left, and bottom-right values of a 2×2 patch, respectively.

$$\begin{cases} LL = \dfrac{a+b+c+d}{2}, \\ LH = \dfrac{-a-b+c+d}{2}, \\ HL = \dfrac{-a+b-c+d}{2}, \\ HH = \dfrac{a-b-c+d}{2}. \end{cases} \tag{15}$$

We collect the three high-frequency subbands as

$$D = [LH, HL, HH], \quad (LL, D) = \mathcal{W}(I_c). \tag{16}$$

Here, $LL$ mainly represents low-frequency color and large-scale structure, while $D$ retains directional edges and local detail.

#### *2) Confidence-Gated Reconstruction*

Adaptive average pooling $pool(\cdot)$ reduces the full-resolution backbone feature to the wavelet-subband resolution:

$$F_w = pool(F). \tag{17}$$

Three independent 3×3 convolutional heads predict low-frequency residuals, high-frequency residuals, and surrogate gate scores, respectively:

$$\begin{cases} \Delta LL = \mathcal{H}_L(F_w), \\ \Delta D = \mathcal{H}_D(F_w), \\ U = clip_{[-6,6]}(\mathcal{H}_U(F_w)). \end{cases} \tag{18}$$

where $\Delta LL \in \mathbb{R}^{3\times h\times w}$ and $D, \Delta D, U, Q \in \mathbb{R}^{3\times3\times h\times w}$. We arrange the high-frequency residuals and gate scores along the RGB-channel and wavelet-direction axes. The predicted scores are clipped to [-6,6] to obtain $U$, which controls the strength of each high-frequency correction. The corresponding confidence $Q$ is

$$Q = \sigma(-U). \tag{19}$$

where $\sigma$ denotes the sigmoid function. Clipping limits $Q$ to approximately [0.0025, 0.9975], allowing high-frequency residuals to be strongly suppressed or almost fully retained. Lower gate scores increase $Q$; higher scores decrease $Q$ and suppress the candidate residual at that location. We then update the low- and high-frequency components:

$$\widehat{LL} = LL + \Delta LL, \quad \widehat{D} = D + Q \odot \Delta D. \qquad (20)$$

We apply the low-frequency residual directly and gate the high-frequency residual by confidence. Here, $\Delta D$ is the candidate correction, and $Q$ determines the fraction added to each detail coefficient.

Inverse Haar reconstruction produces the enhanced image, which is cropped to the original spatial size and clipped to [0,1]:

$$\hat{J} = clip_{[0,1]}\left(Crop_{H,W}\left[\mathcal{W}^{-1}\left(\widehat{LL}, \widehat{D}\right)\right]\right). \qquad (21)$$

All three UGWR prediction heads are zero-initialized, giving $\Delta LL = 0$ and $\Delta D = 0$ at initialization. The network therefore initially preserves the MCT output approximately and learns the required wavelet corrections during training.

### *E. Reference-Anchored Degradation Replay*

Paired supervision encourages the enhanced image to approximate its reference, but does not directly test consistency with the degraded observation. Re-degrading the enhanced image with a learned renderer provides an additional constraint. Joint optimization may allow the restoration network and renderer to partially compensate for each other. RADR is designed to reduce this ambiguity by rendering the enhanced image and its paired reference with the same degradation parameters. A predictor maps the bottleneck feature to latent attenuation, background-light, and range proxies. We then use the observed input as the reconstruction target for both rendered images. The reference path constrains the renderer through the paired target, and the prediction path ties the enhanced output to the same estimated degradation.

#### *1) Shared Degradation Estimation*

Given the bottleneck feature $Z$, a global branch predicts 6-dimensional water parameters and a spatial branch estimates the distance proxy map. The global prediction head computes

$$\omega = \mathcal{H}_{water}\left(GAP(Z)\right). \qquad (22)$$

Here, $GAP$ represents global average pooling over the spatial dimensions. Softplus makes the three channel-wise attenuation parameters positive, after which we normalize them by their mean:

$$\beta = \frac{Softplus(\omega_{1:3})}{Mean[Softplus(\omega_{1:3})]}. \qquad (23)$$

Here, $Mean$ averages the three channel-wise attenuation values. We define the background light and distance proxy as

$$A = \sigma(\omega_{4:6}), \quad d = Resize_{H,W}\left[\sigma\left(\mathcal{H}_d(Z)\right)\right]. \qquad (24)$$

The operator $\mathcal{H}_d$ represents the spatial range-prediction head. $\beta$ and $A$ are image-level three-channel vectors, and $d$ is a single-channel spatial map. Resize denotes bilinear interpolation to the input resolution. Thus, $\beta$ represents relative attenuation across the three RGB channels, while $d$ acts as a spatially varying distance/degradation proxy. These variables are learned through reconstruction consistency and parameterize the replay model. They do not represent independently measured optical properties or metric scene depth.

#### *2) Reference-Anchored Replay*

These parameters define the transmittance:

$$t_c(x, y) = \exp\left(-\beta_c d(x, y)\right). \qquad (25)$$

where $c$ indexes the RGB channel and $(x, y)$ denotes a spatial location. The range map d is shared across channels. A simplified underwater image formation model then re-renders the image:

$$\mathcal{R}(X) = X \odot t + A \odot (1 - t). \qquad (26)$$

Both the enhanced image $\hat{J}$ and reference image $J$ are rendered with the same parameter set $\{\beta, A, d\}$, giving two replay paths:

$$\tilde{I}_p = \mathcal{R}(\hat{J}; \beta, A, d), \quad \tilde{I}_r = \mathcal{R}(J; \beta, A, d). \qquad (27)$$

Reference replay anchors the degradation predictor to the paired observation; prediction replay links the enhanced output to the same parameters. Both terms update the predictor and its upstream features. The two replay paths therefore share the same degradation parameters. RADR provides auxiliary consistency supervision during training and is omitted at inference.

### *F. Training Objective*

The training objective combines output fidelity, UGWR confidence calibration, and RADR replay consistency.

#### *1) Fidelity and Replay Objectives*

We combine Charbonnier loss and SSIM to constrain pixel reconstruction and local structure:

$$\mathcal{L}_{\text{fid}} = \mathcal{L}_{\text{char}}(\hat{J}, J) + \lambda_s\left[1 - \text{SSIM}(\hat{J}, J)\right]. \qquad (28)$$

where $\lambda_s = 0.2$. We clip the SSIM score in the fidelity loss to [0,1] and define the Charbonnier loss as

$$\mathcal{L}_{\text{char}}(X, Y) = \frac{1}{N}\sum_{i=1}^{N}\sqrt{(X_i - Y_i)^2 + \varepsilon^2}, \quad \varepsilon = 10^{-3}. \qquad (29)$$

where $N$ is the total number of image elements. RADR requires both the prediction and reference replay paths to reconstruct the original observation $I$:

$$\mathcal{L}_{\text{rep}} = \frac{1}{2}\left[\mathcal{L}_{\text{char}}(\tilde{I}_p, I) + \mathcal{L}_{\text{char}}(\tilde{I}_r, I)\right]. \qquad (30)$$

We give the two terms equal weights, allowing prediction replay and the reference anchor to optimize the degradation variables jointly.

#### *2) Confidence Calibration*

Eq. (19) bounds $Q$, but the sigmoid parameterization alone does not define a reconstruction-reliability target for the gate. We calibrate the gate against the wavelet-detail reconstruction error relative to the reference image. Under this definition, higher confidence corresponds to smaller errors relative to the adaptive normalization scale.

TABLE I
QUANTITATIVE COMPARISON OF VARIOUS UIE METHODS ON FOUR FULL-REFERENCE TESTING DATASETS. THE BEST, SECOND-BEST, AND THIRD-BEST RESULTS ARE HIGHLIGHTED IN RED, BLUE, AND GREEN, RESPECTIVELY.

| Methods | Test-E515 | | | | Test-L400 | | | | Test-U90 | | | | Test-UFO120 | | | |
|---|---|---|---|---|---|---|---|---|---|---|---|---|---|---|---|---|
| | PSNR↑ | SSIM↑ | LPIPS↓ | FID↓ | PSNR↑ | SSIM↑ | LPIPS↓ | FID↓ | PSNR↑ | SSIM↑ | LPIPS↓ | FID↓ | PSNR↑ | SSIM↑ | LPIPS↓ | FID↓ |
| UWCNN (2020 PR) | 17.27 | 0.732 | 0.381 | 68.57 | 16.95 | 0.744 | 0.397 | 75.58 | 14.94 | 0.699 | 0.334 | 103.09 | 17.11 | 0.701 | 0.383 | 94.82 |
| WaterNet (2020 TIP) | 20.25 | 0.805 | 0.276 | 33.42 | 21.91 | 0.855 | 0.237 | 40.47 | 22.78 | 0.881 | 0.113 | 29.76 | 19.70 | 0.772 | 0.314 | 55.59 |
| Ucolor (2021 TIP) | 19.98 | 0.813 | 0.272 | 34.14 | 20.55 | 0.835 | 0.256 | 41.61 | 22.34 | 0.868 | 0.118 | 33.66 | 19.95 | 0.780 | 0.311 | 53.80 |
| SGUIE-Net (2022 TIP) | 18.65 | 0.773 | 0.302 | 38.64 | 20.70 | 0.824 | 0.239 | 43.42 | 23.52 | 0.900 | 0.106 | 31.03 | 19.17 | 0.763 | 0.288 | 54.42 |
| TACL (2022 TIP) | 18.93 | 0.767 | 0.264 | 36.57 | 20.64 | 0.835 | 0.222 | 40.64 | 23.41 | 0.883 | 0.114 | 31.72 | 19.14 | 0.760 | 0.271 | 54.35 |
| U-shape (2023 TIP) | 27.78 | 0.861 | 0.176 | 28.74 | 23.94 | 0.836 | 0.197 | 46.06 | 20.51 | 0.778 | 0.205 | 65.24 | 23.44 | 0.799 | 0.221 | 54.14 |
| TUDA (2023 TIP) | 16.84 | 0.754 | 0.340 | 53.57 | 18.28 | 0.809 | 0.288 | 63.30 | 18.93 | 0.856 | 0.193 | 72.58 | 16.12 | 0.714 | 0.370 | 79.31 |
| NU²Net (2023 AAAI) | 19.50 | 0.800 | 0.290 | 37.41 | 21.35 | 0.848 | 0.243 | 43.87 | 23.08 | 0.904 | 0.114 | 31.23 | 24.39 | 0.816 | 0.273 | 55.47 |
| P2CNet (2024 TCSVT) | 17.47 | 0.747 | 0.332 | 43.03 | 17.90 | 0.774 | 0.299 | 51.26 | 19.51 | 0.823 | 0.193 | 55.23 | 17.48 | 0.725 | 0.341 | 68.47 |
| CCL-Net (2025 TMM) | 24.57 | 0.822 | 0.161 | 33.08 | 24.84 | 0.872 | 0.142 | 35.51 | 21.80 | 0.886 | 0.143 | 40.65 | 24.85 | 0.814 | 0.139 | 50.93 |
| SS-UIE (2025 AAAI) | 26.56 | 0.858 | 0.188 | 27.04 | 27.52 | 0.878 | 0.163 | 39.67 | 21.36 | 0.858 | 0.156 | 44.32 | 26.23 | 0.842 | 0.174 | 41.61 |
| PSPL (2025 TIP) | 16.36 | 0.666 | 0.373 | 82.24 | 18.27 | 0.743 | 0.312 | 76.43 | 19.10 | 0.821 | 0.237 | 75.74 | 16.62 | 0.660 | 0.341 | 68.47 |
| GHS-UIR (2025 TCSVT) | 24.63 | 0.849 | 0.149 | 30.18 | 25.45 | 0.872 | 0.119 | 38.73 | 23.48 | 0.893 | 0.109 | 30.50 | 24.54 | 0.818 | 0.149 | 52.64 |
| SUCode (2026 TGRS) | 18.69 | 0.795 | 0.271 | 35.08 | 19.86 | 0.822 | 0.230 | 41.76 | 18.49 | 0.762 | 0.262 | 56.14 | 21.14 | 0.844 | 0.158 | 39.90 |
| HRMamba (2026 TIP) | 27.11 | 0.868 | 0.192 | 27.38 | 28.04 | 0.909 | 0.139 | 28.92 | 21.85 | 0.878 | 0.126 | 34.34 | 26.93 | 0.844 | 0.204 | 46.84 |
| MCL-UIE (Ours) | 28.53 | 0.891 | 0.143 | 24.26 | 28.09 | 0.917 | 0.112 | 25.23 | 23.37 | 0.911 | 0.100 | 25.16 | 28.02 | 0.868 | 0.148 | 42.08 |

TABLE II
NO-REFERENCE QUALITY AND COMPUTATIONAL EFFICIENCY COMPARISON OF VARIOUS UIE METHODS. UCCS (AVG.) DENOTES THE AVERAGE PERFORMANCE OVER THE UCCS-BLUE, UCCS-BLUE-GREEN, AND UCCS-GREEN SUBSETS. THE BEST, SECOND-BEST, AND THIRD-BEST RESULTS ARE HIGHLIGHTED IN RED, BLUE, AND GREEN, RESPECTIVELY.

| Methods | Test-C60 | | Test-U45 | | UCCS (Avg.) | | Efficiency | | |
|---|---|---|---|---|---|---|---|---|---|
| | UIQM↑ | UCIQE↑ | UIQM↑ | UCIQE↑ | UIQM↑ | UCIQE↑ | FLOPs (G) | Params (M) | GPU Runtime (ms) |
| UWCNN (2020 PR) [14] | 2.270 | 0.544 | 2.755 | 0.577 | 2.777 | 0.507 | 2.615 | 0.040 | 2.927 |
| WaterNet (2020 TIP) [15] | 2.697 | 0.580 | 3.187 | 0.590 | 3.099 | 0.569 | 71.452 | 1.091 | 8.827 |
| Ucolor (2021 TIP) [26] | 2.627 | 0.554 | 3.196 | 0.588 | 2.933 | 0.536 | 443.855 | 157.421 | 89.310 |
| SGUIE-Net (2022 TIP) [43] | 2.715 | 0.586 | 3.188 | 0.613 | 3.083 | 0.550 | 70.786 | 1.095 | 7.174 |
| TACL (2022 TIP) [44] | 2.745 | 0.579 | 3.214 | 0.625 | 3.119 | 0.584 | 56.864 | 11.378 | 4.316 |
| U-shape (2023 TIP) [16] | 2.583 | 0.558 | 2.978 | 0.586 | 3.004 | 0.556 | 66.204 | 65.617 | 39.275 |
| TUDA (2023 TIP) [24] | 2.788 | 0.582 | 3.324 | 0.590 | 3.149 | 0.563 | 175.039 | 29.876 | 26.539 |
| NU²Net (2023 AAAI) [45] | 2.761 | 0.571 | 3.243 | 0.593 | 3.056 | 0.551 | 10.486 | 3.146 | 1.550 |
| P2CNet (2024 TCSVT) [46] | 2.769 | 0.598 | 3.320 | 0.621 | 3.085 | 0.577 | 5.180 | 2.068 | 2.840 |
| CCL-Net (2025 TMM) [17] | 2.683 | 0.581 | 3.310 | 0.608 | 3.048 | 0.560 | 108.145 | 0.553 | 6.346 |
| SS-UIE (2025 AAAI) [28] | 2.777 | 0.579 | 3.103 | 0.600 | 2.993 | 0.554 | 46.764 | 20.634 | 51.291 |
| PSPL (2025 TIP) [29] | 2.752 | 0.586 | 3.425 | 0.597 | 3.194 | 0.569 | 0.976 | 0.101 | 0.923 |
| GHS-UIR (2025 TCSVT) [47] | 2.599 | 0.574 | 3.184 | 0.603 | 2.979 | 0.553 | 10.053 | 1.145 | 55.628 |
| SUCode (2026 TGRS) [48] | 2.661 | 0.568 | 3.206 | 0.589 | 3.013 | 0.555 | 333.758 | 68.220 | 93.563 |
| HRMamba (2026 TIP) [49] | 2.651 | 0.566 | 3.049 | 0.592 | 3.013 | 0.564 | 0.594 | 1.067 | 15.875 |
| MCL-UIE (Ours) | 2.521 | 0.571 | 3.111 | 0.602 | 2.998 | 0.565 | 10.098 | 2.487 | 6.278 |

Applying the Haar transform to the reference image $J$ gives its detail coefficients $D_J$. We define the current high-frequency reconstruction error as

$$E = \left|\widehat{D} - D_J\right|. \tag{31}$$

We detach the error before constructing the calibration target. For each image $n$ and color channel $c$, the adaptive error scale averages this detached error over the three wavelet directions $k$ and all spatial locations, with a lower bound of $10^{-3}$:

$$\tau_{n,c} = \max\left(Mean_{k,x,y}\left[sg\left(E_{n,c,k,x,y}\right)\right], 10^{-3}\right),$$
$$Q^* = \exp\left(-\frac{\mathrm{sg}(E)}{\tau}\right). \tag{32}$$

where $Q^*$ is the calibration target and $\tau_{n,c}$ is broadcast over wavelet directions and spatial locations. $sg(\cdot)$ denotes stop-gradient; $\tau$ is a data-dependent normalization scale, not a learnable parameter. We calibrate the predicted confidence with Smooth L1 loss:

$$\mathcal{L}_{\mathrm{cal}} = \mathrm{SmoothL1}(Q, Q^*). \tag{33}$$

Smaller normalized errors $E/\tau$ yield higher target confidence. The calibration loss trains the gate to follow this error-based target, while the fidelity loss optimizes the enhanced image. Stop-gradient blocks gradients from the calibration target from propagating through the reconstructed detail coefficients. Confidence is therefore a learned weight calibrated to normalized reconstruction error, rather than a probability that a detail coefficient is correct. The objective ablation tests how calibration affects final image quality, both with and without replay supervision.

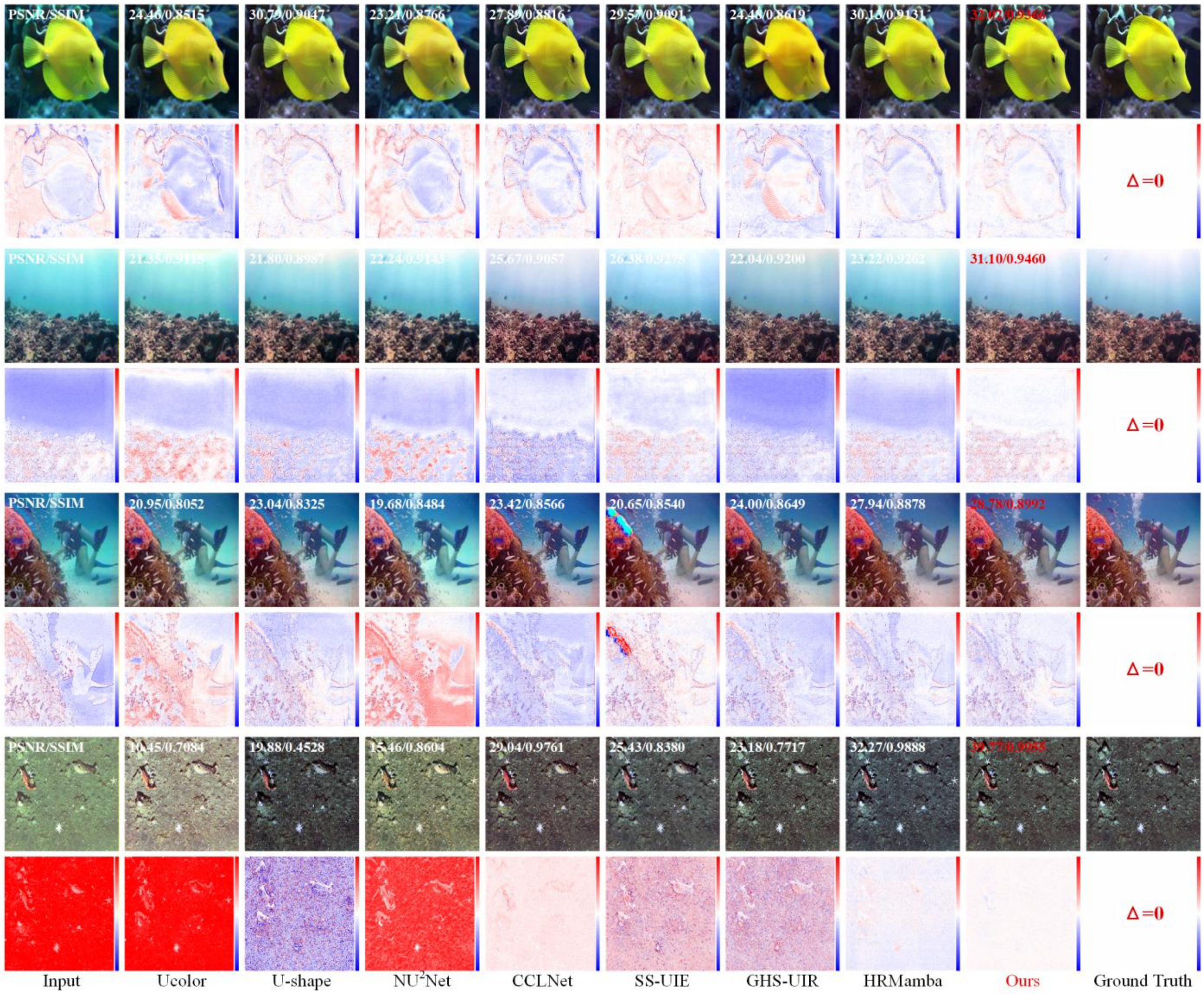

Fig. 4. Full-reference qualitative comparisons on Test-E515, Test-L400, Test-U90, and Test-UFO120. Each representative example is accompanied by an error map relative to the ground truth to visualize restoration deviations.

*3) Overall Objective*

The final objective combines fidelity, calibration, and replay:

$$\mathcal{L}_{\text{total}} = \mathcal{L}_{\text{fid}} + 0.1(\mathcal{L}_{\text{rep}} + \mathcal{L}_{\text{cal}}). \quad (34)$$

## IV. Experiments

### *A. Experimental Details*

We implement MCL-UIE in PyTorch and train it on an NVIDIA GeForce RTX 4090D GPU using AdamW, with an initial learning rate of $2\times10^{-4}$ and weight decay of $1\times10^{-4}$. Training runs for 300 epochs with a batch size of 8, and cosine annealing reduces the learning rate to $2\times10^{-6}$. We apply the same augmentations to each input-reference pair: random cropping to 256×256, horizontal and vertical flips, and rotations of 90, 180, and 270 degrees. Images smaller than the crop size are resized by bicubic interpolation.

### *B. Experimental Settings*

*1) Datasets*

We train on two publicly available paired datasets: UIEB [15] and LSUI [16]. UIEB contains 890 underwater images with corresponding references. Following the commonly used split, we use 800 pairs for training and 90 for testing. LSUI contains 4279 input-reference pairs, of which 3879 are used for training and 400 for testing.

We evaluate generalization across underwater environments using paired and no-reference datasets. The paired test sets contain 515 image pairs from EUVP [23], 400 from LSUI, 90 from UIEB, and 120 from UFO-120 [40], denoted as Test-E515, Test-L400, Test-U90, and Test-UFO120, respectively. The no-reference sets comprise 60 challenge images from UIEB and 45 images from U45 [41], denoted as Test-C60 and Test-U45. We also select 300 images from the UCCS subset of RUIE [42]: 100 each from UCCS-blue, UCCS-green, and UCCS-blue-green, forming Test-UCCS-blue, Test-UCCS-green, and Test-UCCS-blue-green. These datasets cover

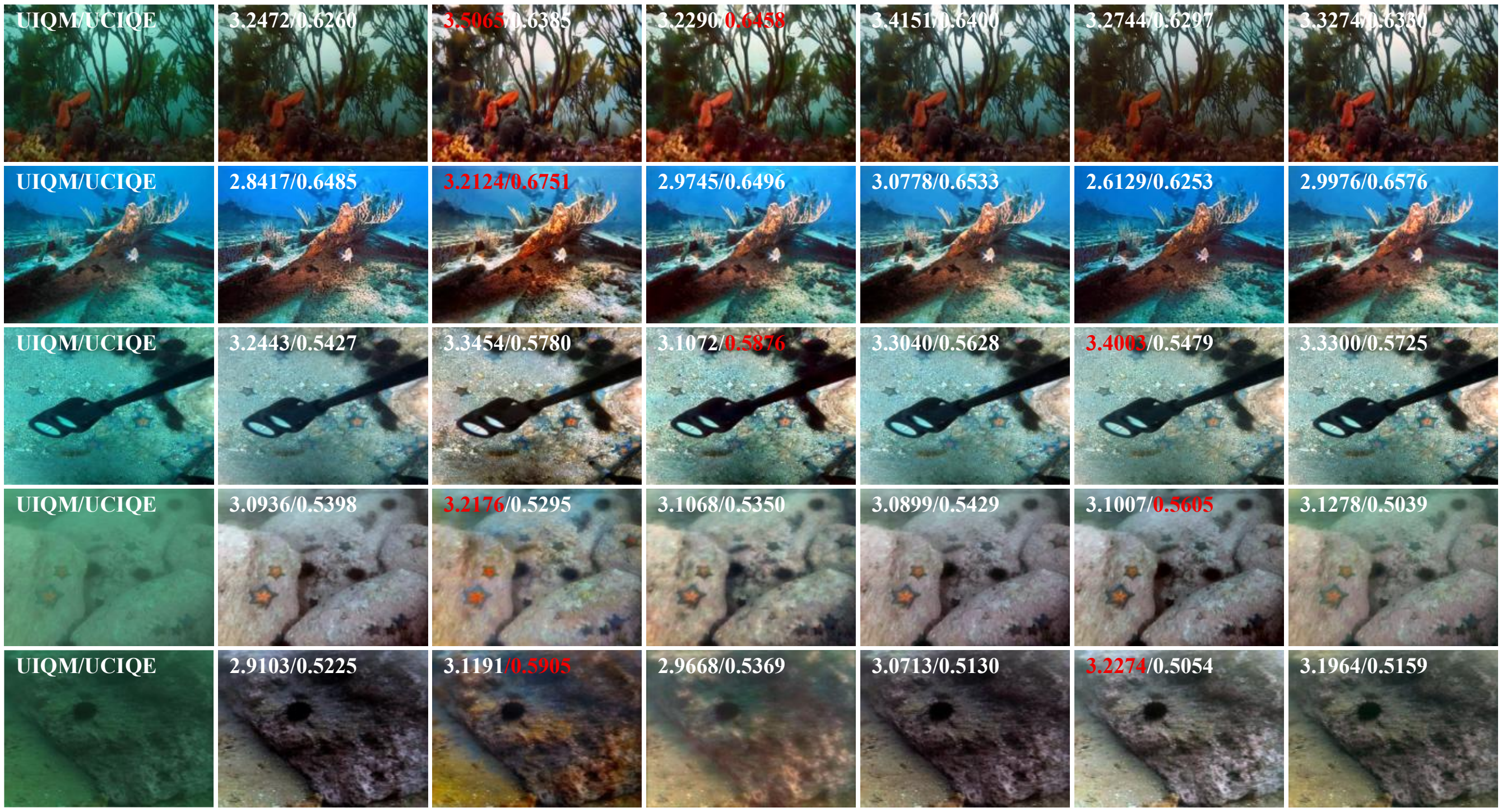

Fig. 5. No-reference qualitative comparisons on Test-C60, Test-U45, and the three RUIE-UCCS subsets (blue, blue-green, and green). The examples illustrate the generalization of MCL-UIE under diverse color casts and real-world underwater degradations.

TABLE III
COMPONENT-WISE ABLATION ANALYSIS OF MCT, UGWR, AND RADR ON THE TEST-E515, TEST-L400, TEST-U90, AND TEST-UFO120 DATASETS.

| Model | | | | Test-E515 | | | Test-L400 | | | Test-U90 | | | Test-UFO120 | | |
|---|---|---|---|---|---|---|---|---|---|---|---|---|---|---|---|
| Baseline | MCT | UGWR | RADR | PSNR↑ | SSIM↑ | LPIPS↓ | PSNR↑ | SSIM↑ | LPIPS↓ | PSNR↑ | SSIM↑ | LPIPS↓ | PSNR↑ | SSIM↑ | LPIPS↓ |
| √ | × | × | × | 27.64 | 0.885 | 0.155 | 26.85 | 0.909 | 0.126 | 22.25 | 0.902 | 0.116 | 26.78 | 0.859 | 0.167 |
| √ | √ | × | × | 28.14 | 0.890 | 0.149 | 27.15 | 0.912 | 0.120 | 23.01 | 0.906 | 0.106 | 27.29 | 0.859 | 0.159 |
| √ | √ | √ | × | 28.34 | 0.889 | 0.150 | 27.32 | 0.912 | 0.121 | 23.15 | 0.908 | 0.104 | 27.71 | 0.863 | 0.155 |
| √ | √ | √ | √ | 28.53 | 0.891 | 0.143 | 28.09 | 0.917 | 0.112 | 23.37 | 0.911 | 0.100 | 28.02 | 0.868 | 0.148 |

varied water conditions, color casts, and degradation levels for testing generalization.

For a compact quantitative comparison, Table II averages the results on UCCS-blue, UCCS-blue-green, and UCCS-green under UCCS (Avg.). The qualitative comparison presents representative samples from each subset separately.

*2) Comparative Methods*

We compare MCL-UIE with fifteen UIE methods covering different architectures and learning paradigms: UWCNN [14], WaterNet [15], Ucolor [26], SGUIE-Net [43], TACL [44], U-shape [16], TUDA [24], $NU^2$Net [45], P2CNet [46], CCL-Net [17], SS-UIE [28], PSPL [29], GHS-UIR [47], SUCode [48], and HRMamba [49]. We use public implementations and official pretrained weights with their original inference settings, while applying a common preprocessing and evaluation protocol to improve comparability across methods.

*3) Evaluation Metrics*

For paired datasets, PSNR measures pixel fidelity, SSIM [50] measures structural fidelity, and LPIPS [51] measures deep perceptual distance. FID [52] evaluates distribution-level similarity to the reference set. For no-reference datasets, we report UIQM [53] and UCIQE [54] as complementary indicators of color, contrast, and sharpness. These metrics do not fully capture perceptual naturalness.

## *C. Quantitative Results*

*1) Full-Reference Comparison*

Table I compares full-reference results on four paired datasets. MCL-UIE ranks first in 13 of the 16 dataset-metric combinations, including all four metrics on Test-E515 and Test-L400. Compared with the second-best method, it improves PSNR by 0.75 dB on Test-E515 and reduces FID by 3.69 on Test-L400. On Test-U90, it achieves the best SSIM, LPIPS, and FID. Its PSNR of 23.37 dB is 0.15 dB below the best result. On Test-UFO120, it ranks first in PSNR and SSIM, second in LPIPS, and third in FID. These gains span pixel, structural, perceptual, and distribution-level measures, although rankings vary by dataset and metric. Test-UFO120 illustrates this distinction: the best pixel and structural fidelity need not coincide with the best perceptual distance or distributional match.

*2) No-Reference Comparison*

Table II reports no-reference results on Test-C60 and Test-U45, along with averages over UCCS-blue, UCCS-blue-green,

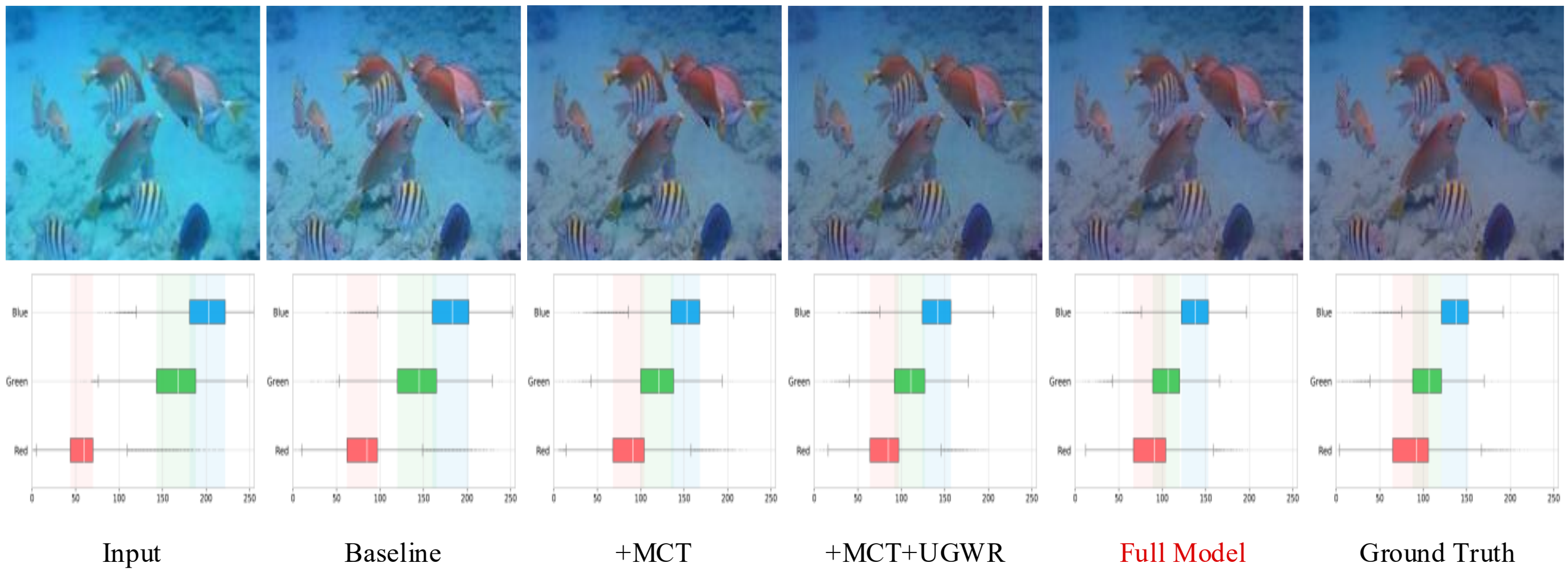

Fig. 6. Component-wise qualitative ablation and RGB distribution analysis. Outputs of the baseline and progressive MCL-UIE variants are compared with the ground truth to illustrate the effects of MCT, UGWR, and RADR on restoration quality and global color distribution.

and UCCS-green. MCL-UIE is competitive across these conditions, though it does not consistently achieve the highest handcrafted metric scores. It obtains a UCIQE of 0.602 on Test-U45 and an average UIQM/UCIQE of 2.998/0.565 across the three UCCS subsets. UIQM and UCIQE emphasize colorfulness, contrast, and sharpness, so stronger enhancement can increase the scores without improving naturalness. We interpret these results alongside the visual comparisons. They support performance across the evaluated color casts, but the relative quality of individual outputs depends on the metric and visual criterion used.

### *D. Model Complexity Analysis*

Table II also compares FLOPs, parameter counts, and GPU inference latency with representative UIE methods. For a 256 $\times$ 256 input, MCL-UIE has 2.487M parameters, requires 10.098 GFLOPs, and takes 6.278 ms on an RTX 4090D GPU. Several lightweight methods cost less to run. At its reported computational cost, MCL-UIE achieves the highest PSNR and SSIM on Test-E515, Test-L400, and Test-UFO120, and the best SSIM, LPIPS, and FID on Test-U90. These results describe the trade-off between restoration fidelity and computation. RADR contributes its consistency constraint during training and is absent at inference, so the reported costs cover only the restoration path.

### *E. Qualitative Results*

#### *1) Full-Reference Comparison*

Fig. 4 presents one representative example from each paired test set with the corresponding error maps. Deeper red or blue indicates a larger positive or negative deviation from the reference images. In the selected examples, MCL-UIE exhibits relatively small color deviation, and its error map is closer to white than those of other methods. It also achieves the highest PSNR and SSIM in all four examples. On Test-E515 and Test-L400, it restores fish and coral colors while preserving a natural water background and balanced lighting. On Test-U90, it retains the contours and fine textures of the diver and fish without noticeable color artifacts. On Test-UFO120, it recovers foreground objects and low-contrast background detail, producing results close to the reference. Competing methods show residual blue or green casts, excessive red enhancement, uneven brightness, or lost local detail. The examples show improvements in both global color and local structure. Component ablations examine the contribution of each constraint to these changes.

#### *2) No-Reference Comparison*

Fig. 5 compares results on Test-C60, Test-U45, and representative samples from UCCS-blue, UCCS-blue-green, and UCCS-green. We show the UCCS subsets separately to examine different color casts, although their scores are averaged in Table II. MCL-UIE improves visibility in the kelp, shipwreck, starfish, and seabed scenes while retaining relatively natural color and contrast. Some competing methods produce stronger saturation or contrast and higher UIQM or UCIQE scores, but also introduce warm color biases or unnatural local tones in these examples. MCL-UIE yields more restrained results across blue, green, and mixed casts. These comparisons show where higher metric scores can diverge from visual naturalness. Because paired references are unavailable, the observations describe visible restoration behavior rather than fidelity to an unknown clean image.

### *F. Ablation Study*

We evaluate MCT, UGWR, and RADR through progressive component-wise ablations. The baseline in Table III uses the same U-shaped restoration backbone with all three constraints disabled. Sequentially adding MCT, UGWR, and RADR gives four configurations, trained under identical settings and evaluated on the same four test sets. Each comparison measures the gain over the preceding configuration. The backbone remains fixed throughout, but each measured gain depends on the components already present. We also ablate the training objectives to examine the separate and combined effects of calibration and replay.

#### *1) Effectiveness of MCT*

Adding MCT to the baseline improves PSNR by 0.30 to 0.76 dB across the four test sets, with an average gain of 0.52

TABLE IV
ABLATION OF AUXILIARY TRAINING OBJECTIVES ON FOUR FULL-REFERENCE TEST SETS. THE FIDELITY LOSS IS RETAINED IN ALL VARIANTS.

| Model | | | Test-E515 | | | Test-L400 | | | Test-U90 | | | Test-UFO120 | | |
|---|---|---|---|---|---|---|---|---|---|---|---|---|---|---|
| $L_{fid}$ | $L_{cal}$ | $L_{rep}$ | PSNR↑ | SSIM↑ | LPIPS↓ | PSNR↑ | SSIM↑ | LPIPS↓ | PSNR↑ | SSIM↑ | LPIPS↓ | PSNR↑ | SSIM↑ | LPIPS↓ |
| √ | × | × | 27.93 | 0.887 | 0.149 | 28.04 | 0.916 | 0.109 | 23.35 | 0.911 | 0.100 | 27.71 | 0.865 | 0.151 |
| √ | √ | × | 26.95 | 0.878 | 0.163 | 27.59 | 0.913 | 0.117 | 23.17 | 0.907 | 0.103 | 27.03 | 0.858 | 0.161 |
| √ | × | √ | 28.47 | 0.891 | 0.144 | 27.88 | 0.917 | 0.109 | 23.26 | 0.911 | 0.101 | 27.97 | 0.867 | 0.149 |
| √ | √ | √ | 28.53 | 0.891 | 0.143 | 28.09 | 0.917 | 0.112 | 23.37 | 0.911 | 0.100 | 28.02 | 0.868 | 0.148 |

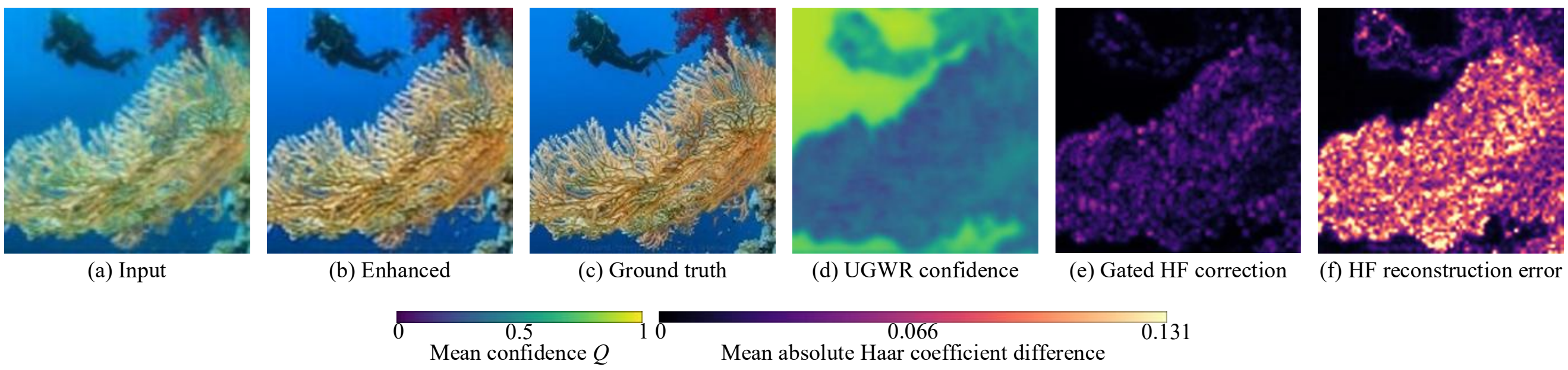

(a) Input (b) Enhanced (c) Ground truth (d) UGWR confidence (e) Gated HF correction (f) HF reconstruction error

Fig. 7. UGWR confidence-gated high-frequency reconstruction in a selected Test-L400 crop. (a–c) Input, enhanced output, and paired reference. (d) Mean confidence $Q$. (e) Mean magnitude of the applied high-frequency correction $|Q \odot \Delta D|$. (f) Mean absolute wavelet-detail error relative to the reference. Maps are averaged over RGB channels and Haar directions.

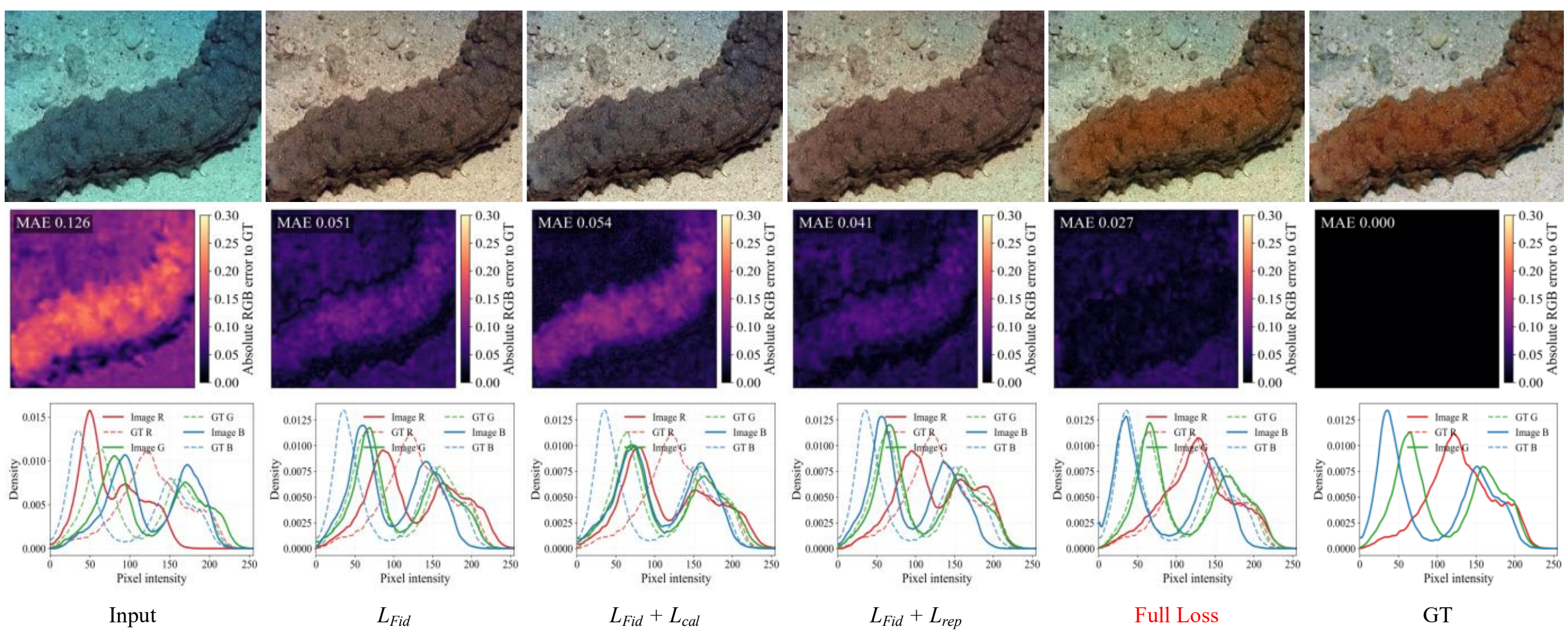

Fig. 8. Qualitative analysis of different training objectives on a representative LSUI sample. The first row shows restoration results under different objective configurations, the second row presents absolute RGB error maps relative to the ground truth, and the third row compares the corresponding RGB intensity distributions.

dB. SSIM increases by 0.003 on average, and LPIPS decreases by 0.008. In Fig. 6, MCT shifts the RGB intensity distributions closer to those of the reference images and reduces the channel imbalance in the baseline outputs. These distribution changes agree with the reconstruction gains, supporting explicit global color correction before spatial restoration. The gains apply to MCT as a complete module, including its statistical descriptor, triangular parameterization, and coefficient bounds.

*2) Effectiveness of UGWR*

Adding UGWR to the MCT configuration further improves PSNR on all four test sets by 0.14 to 0.42 dB. The largest improvement occurs on Test-UFO120, where SSIM rises from 0.859 to 0.863 and LPIPS falls from 0.159 to 0.155. The RGB box plots in Fig. 6 change only moderately with UGWR. This pattern is consistent with better local reconstruction and little change in global color statistics. However, marginal RGB distributions cannot locate the wavelet corrections. The local visualizations therefore give a more direct view of where UGWR changes the image.

Fig. 7 illustrates UGWR's spatial response in a Test-L400 crop. Compared with the input, the enhanced image renders the coral-like structures more clearly and appears closer to the paired reference. The confidence map varies across the crop, while the gated high-frequency correction is concentrated around textured structures. The wavelet-detail error map shows where discrepancies from the reference remain.

*3) Overall Effect and Qualitative Analysis*

The complete MCL-UIE configuration yields the strongest aggregate performance among the evaluated variants in Table

III. Relative to the baseline, it improves PSNR by 1.12 dB and SSIM by 0.008 on average, while reducing LPIPS by 0.015. Fig. 6 shows natural tones and color distributions close to the references, without noticeable oversaturation. Along the evaluated sequence, global color correction improves the backbone input, wavelet reconstruction adds local corrections, and degradation replay provides additional training supervision. Their combination yields the strongest overall results among the evaluated variants, suggesting that the three constraints are complementary within this architecture.

#### 4) *Effectiveness of Different Training Objectives*

To assess the training objectives, we retain the fidelity loss in every variant and add confidence calibration and degradation replay separately or together. Table IV shows that calibration alone reduces performance under the evaluated settings, whereas replay provides clear gains on Test-E515 and Test-UFO120 and mixed effects on Test-L400 and Test-U90. Combining both objectives yields the highest PSNR on all four datasets: 28.53, 28.09, 23.37, and 28.02 dB on Test-E515, Test-L400, Test-U90, and Test-UFO120, respectively. The combination also achieves the best or joint-best SSIM. LPIPS increases slightly on Test-L400, but reaches the best or joint-best value on the other datasets. Calibration improves reconstruction when combined with replay. Used alone, it reduces performance under this protocol. The two objectives constrain different aspects of restoration: calibration trains a local error-based gate, while replay enforces consistency with the observed image through shared degradation parameters. Their combined improvement supports supervising both local corrections and observation consistency.

Fig. 8 compares the training objectives on a representative Test-L400 sample. Replay supervision reduces RGB MAE from 0.051 to 0.041, while calibration alone raises it to 0.054. The complete objective gives the lowest MAE of 0.027, accompanied by a lower-intensity error map and an RGB distribution closer to the reference images. The sample follows the dataset-level pattern, in which replay improves reconstruction on its own and calibration provides an additional benefit when combined with replay. The error map and RGB distribution show how this quantitative improvement appears in the reconstructed image.

## V. Conclusion

We present MCL-UIE, a multi-level constrained framework for underwater image enhancement that extends paired-reference supervision from the final output to intermediate restoration operations and degradation consistency. The proposed MCT constrains global chromatic transformation, UGWR regulates local high-frequency reconstruction through calibrated confidence, and RADR introduces reference-anchored degradation consistency during training. Together, these components provide complementary constraints on global color correction, local detail restoration, and the relation between the enhanced result and the observed underwater image. Experiments on paired and no-reference benchmarks demonstrate competitive restoration performance with a compact inference architecture, while progressive component and objective ablations indicate complementary contributions from the three constraints. In particular, the results suggest that confidence calibration is more effective when combined with replay supervision, highlighting the interaction between local correction control and observation-level consistency. Since RADR is used only during training, the additional degradation-consistency supervision does not increase inference-time computation. MCL-UIE currently relies on paired references during training, and the latent attenuation, background-light, and range variables in RADR are learned consistency proxies rather than directly measured physical quantities. Future work may investigate weaker supervision, more flexible degradation models, and adaptation to diverse real-world underwater environments.